\documentclass[letterpaper,10pt,conference]{ieeeconf}
\IEEEoverridecommandlockouts
\usepackage[T1]{fontenc}
\usepackage{amsmath,amssymb}
\usepackage{graphicx}
\usepackage{booktabs,array,tabularx}
\usepackage[table]{xcolor}
\usepackage{cite}
\usepackage{url}
\usepackage{capt-of}
\usepackage{etoc}
\usepackage{ragged2e}
\makeatletter
\let\original@makecaption\@makecaption
\long\def\@makecaption#1#2{%
  \ifx\@captype\@IEEEtablestring
    {\footnotesize\normalfont\justifying\noindent #1: #2\par}%
    \vskip 6pt
  \else
    \original@makecaption{#1}{#2}%
  \fi
}
\makeatother

\usepackage{flushend}
\makeatletter
\let\inputtable\@@input
\makeatother
\definecolor{paperblue}{HTML}{087F8C}
\definecolor{paperbluefill}{HTML}{EDF5F5}
\definecolor{paperslate}{HTML}{647780}

\newcommand{\parahead}[1]{\par\vspace{0.2em}\noindent\textbf{#1}}

\newcommand{\ssecv}{\vspace{-0.1em}}

\definecolor{darkblue}{HTML}{00B7CD}
\definecolor{darkgreen}{HTML}{2E7D32}

\title{\LARGE \bf Reconstruct, Practice, Go Real:\\Guided Self-Improvement for Embodied Agents}
\vspace{-1em}
\author{
Yen-Jen Wang$^{*,\ddagger,1}$,
Haozhe Jiang$^{*,1}$,
Shuying Deng$^{*,1}$,\\
Haoru Xue$^{\ddagger,1}$,
Weirui Ye$^{3}$,
Rocky Duan$^{2}$,
Nika Haghtalab$^{1}$,
S. Shankar Sastry$^{1}$, \\
Pieter Abbeel$^{\dagger,1,2}$,
Haozhi Qi$^{\dagger,2, 4}$
\vspace{0.5em}\\
$^1$UC Berkeley \quad
$^2$Amazon FAR \quad
$^3$MIT \quad
$^4$University of Chicago
\thanks{${}^*$ Equal contribution. ${}^\dagger$ Equal advising. ${}^\ddagger$ Yen-Jen Wang and Haoru Xue were interns at Amazon FAR during this work.}
\thanks{Project website: \protect\url{https://rpg-robot.github.io}}
}

\IEEEaftertitletext{%
  \begin{minipage}{\textwidth}
    \centering
    \vspace{-1.3em}
    \includegraphics[width=\textwidth]{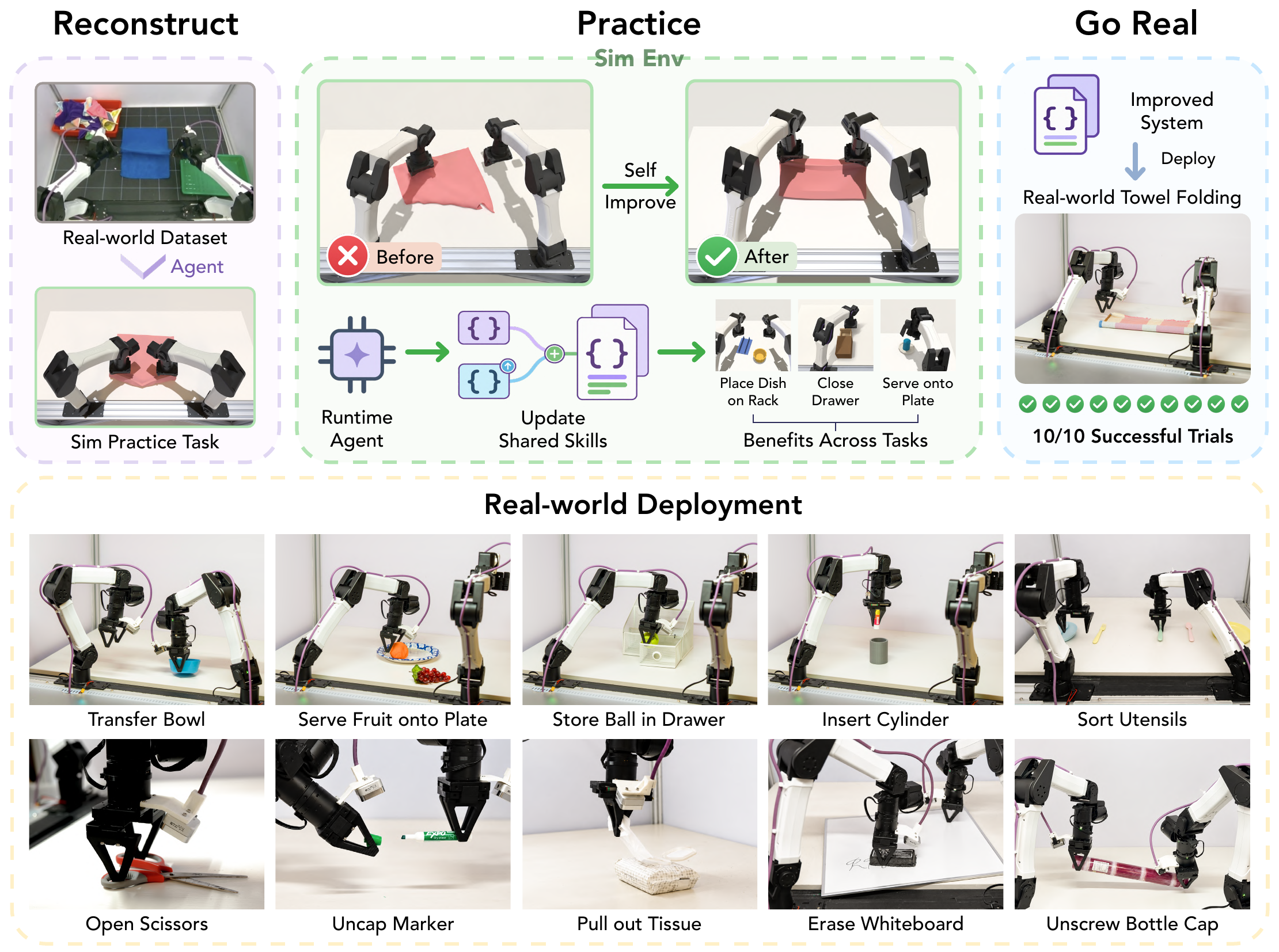}
    \vspace{-7mm}
    \captionof{figure}{\textbf{RPG constructs simulation practice tasks from an offline dataset to develop and refine reusable robot skills.} In \emph{Reconstruct}, an agent identifies manipulation capabilities from an offline dataset and constructs related practice tasks in simulation. In \emph{Practice}, agents use execution feedback to revise shared skills and the system prompt, then validate the changes across tasks before retaining them. Skill revisions can benefit other tasks that use the same library. In \emph{Go Real}, the improved system is calibrated and frozen for deployment on the physical robot. The bottom panel shows qualitative examples of real-world deployment. Quantitative evaluation is conducted on the three tasks reported in Sec.~\ref{sec:physical-results}.}
    \label{fig:teaser}
    \vspace{0.2em}
  \end{minipage}
}

\begin{document}
\maketitle
\thispagestyle{empty}
\pagestyle{empty}

\begin{abstract}
Building reliable robot capabilities across diverse tasks requires substantial human effort to develop and maintain skills, design rewards, and integrate perception with control.
We present Reconstruct, Practice, Go Real (RPG), a framework for autonomous improvement of robot execution systems without updating model weights.
RPG identifies manipulation capabilities in an offline dataset and constructs related practice tasks in simulation.
During practice, RPG uses execution feedback, privileged simulator state, and available dataset videos to diagnose failures.
It develops new reusable symbolic skills, refines existing skills, and revises the system prompt based on these diagnoses.
Cross-task evaluation tests individual candidate changes and merged revisions before they are retained for reuse.
At test time, a multimodal LLM uses the resulting system prompt and skill library to coordinate perception and robot control.
On held-out initializations of 22 manipulation tasks, RPG improves task success from 28.6\% after the first practice round to 95.0\% after 15 rounds, outperforming all evaluated baselines, including ASPIRE (75.5\%) and CaP-Agent0 powered by GPT-6 Astra Pro (60.0\%).
After a common calibration and hardware-adaptation procedure, the frozen system succeeds in all 30 physical trials, with ten trials on each of three tasks.
\end{abstract}

\section{Introduction}

Recent advances in multimodal large language models (MLLMs) have enabled agents to perceive environments and reason about tasks~\cite{palme}. Language-model agents can also generate and execute code~\cite{codeact} and improve through feedback~\cite{reflexion}. These capabilities offer a way to expand robot behavior with less task-specific human engineering. In particular, multimodal coding agents can use visual observations and execution feedback to construct, refine, and compose executable skills that robots can retain and reuse across tasks~\cite{lrll,aspire,playful}.

Language-model-based robot control involves a tradeoff between program execution and online reasoning. \emph{Code-as-Policy} combines perception and control primitives into executable programs~\cite{codeaspolicies,progprompt,capx}. These programs can execute efficiently and incorporate feedback loops, but their responses depend on explicit program logic and perception interfaces. For example, a robust symbolic check for grasp success can be difficult to specify, while an MLLM can assess the outcome from images. In \emph{Agent-as-Policy}, the MLLM remains in the control loop, providing flexible visual and semantic reasoning~\cite{palme,aspire}. However, frequent model calls can make low-level decisions slow. A scalable system should combine these strengths by delegating repeatable execution to reusable code and reserving multimodal reasoning for decisions that benefit from visual or semantic judgment.

We introduce \emph{Reconstruct, Practice, Go Real} (RPG), a framework that develops and refines reusable robot skills through autonomous practice without updating model weights (Fig.~\ref{fig:teaser}). RPG improves both the shared skill library and the system prompt that guides a multimodal agent in using it. Changes to these shared components persist across episodes and can benefit other tasks that reuse them.

Given an offline dataset, \emph{Reconstruct} selects source tasks that expose useful manipulation capabilities and constructs related practice tasks in simulation. The resulting practice suite can be expanded by processing additional tasks from the dataset. A practice task may cover part of a source task or an analogous manipulation. It need not reproduce the original trajectory or scene. We use ABC~\cite{abc} in this work because it covers 193 diverse tasks. During \emph{Practice}, RPG compares executions from the Runtime Agent and the Privileged Agent, which uses the same model and skill library but additionally receives simulator state. The Video Analyzer uses these execution records and available dataset videos to diagnose failures. RPG then develops new reusable symbolic skills, refines existing skills, and revises the system prompt. It evaluates candidate changes across the practice tasks and retests merged changes before retaining them. Finally, \emph{Go Real} calibrates and freezes the resulting system for physical deployment, where the Runtime Agent coordinates perception and control through the shared skill library.

On held-out initializations of 22 manipulation tasks, RPG improves task success from 28.6\% after the first practice round to 95.0\% after 15 rounds, outperforming ASPIRE (75.5\%) and CaP-Agent0 powered by GPT-6 Astra Pro (60.0\%). Through comprehensive ablation experiments, we demonstrate the benefits of the Privileged Agent, Video Analyzer, and skill revisions. After a hardware calibration procedure, the frozen RPG system successfully completes all 30 physical trials, with 10 trials for each of the three evaluated tasks. These results show that autonomous practice can develop and refine shared robot skills that support physical deployment. Our contributions are:
\begin{itemize}
    \item We introduce RPG, a framework that autonomously selects manipulation capabilities from an offline dataset and constructs an expandable suite of simulation practice tasks for developing reusable robot skills.
    \item We develop an autonomous practice procedure that uses privileged execution and video-based failure diagnosis to revise shared skills and the system prompt, validating both individual and merged changes across tasks.
    \item We demonstrate 95.0\% success on held-out initializations of 22 simulated manipulation tasks and 100\% success over 30 trials across three physical tasks.
\end{itemize}

\begin{figure*}[!t]
\centering
\includegraphics[width=1\textwidth]{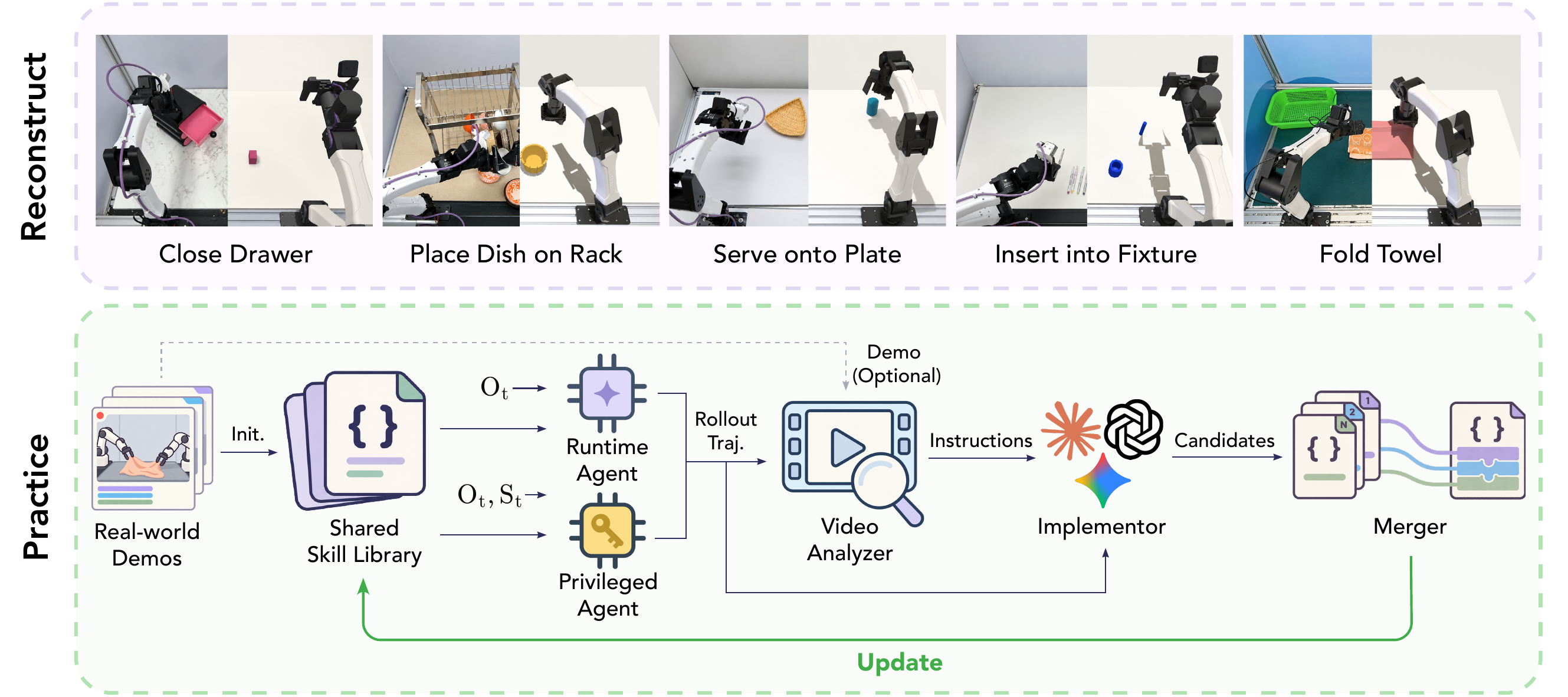}
\vspace{-4mm}
\caption{\textbf{RPG uses privileged execution and video analysis to guide skill development.} The top row shows source scenes alongside representative simulation practice tasks. The Runtime Agent and Privileged Agent share the same model and skill library, but only the latter receives simulator state $S_t$ in addition to robot observations $O_t$. For development tasks, the Video Analyzer compares the two agents' execution records with available videos from the offline dataset to diagnose failures. The Implementor develops new skills, refines existing skills, and revises the system prompt. Candidate changes are evaluated across tasks. The Merger combines selected revisions, and the resulting system is retested before retention.}
\label{fig:architecture}
\vspace{-5.5mm}
\end{figure*}
 
\section{Related Work}

\subsection{Robot Programs and Skill Improvement}

Code as Policies~\cite{codeaspolicies}, ProgPrompt~\cite{progprompt},
and CaP-X~\cite{capx} use language models to generate executable
robot programs, while Prompt2Walk~\cite{wang2023prompt} generates
low-level locomotion commands. SayCan~\cite{saycan} grounds skill
selection in learned affordances, while VoxPoser~\cite{voxposer}
generates 3D value maps for motion planning. Execution feedback also supports
closed-loop reasoning: Inner Monologue~\cite{innermonologue} uses
success detection and scene descriptions to revise plans, and
REFLECT~\cite{reflect} summarizes multisensory execution records to
explain failures and propose corrections.

Agents can retain execution experience for use beyond the current task.
Voyager~\cite{voyager} accumulates executable skills in Minecraft.
Lifelong Robot Library Learning (LRLL)~\cite{lrll} grows a robot skill
library through simulation exploration and code abstraction with human
guidance. ASPIRE~\cite{aspire} diagnoses failures from multimodal traces
and stores validated strategies as in-context skills. RATs~\cite{playful}
uses self-directed play to build reusable skills and failure memories.
RHO~\cite{rho} searches over multi-file robot policy repositories using
execution feedback, paired candidate tests, and validation.
ENPIRE~\cite{enpire} improves robot policies through autonomous
experimentation with physical feedback. 
SHAPER~\cite{sharper} evolves reusable skills and a context-code harness from embodied rollouts without updating model parameters. HERO~\cite{hero} autonomously accumulates and consolidates reusable manipulation behaviors. Outside robotics, GRASP~\cite{grasp} studies regression-gated updates to persistent agent skill libraries.
RPG combines dataset-guided
simulation practice with failure diagnosis that compares executions from the Runtime Agent and Privileged Agent, using source videos when available. It revises
shared skill code and the system prompt, then tests both individual
candidates and merged revisions across a fixed practice suite before
retaining them. RPG performs all self-improvement in simulation and freezes the execution system before physical deployment.

\subsection{Offline Data and Simulation Practice}

Demo2Code~\cite{demo2code} summarizes
state-based demonstrations into task specifications and synthesizes
executable code. Uni-Skill~\cite{uniskill} trains a multimodal planner
and retrieves skill-level video
examples to implement missing skills. RPG uses videos and textual
descriptions from an offline dataset, here ABC~\cite{abc},
to identify manipulation capabilities worth practicing. It constructs
related simulation tasks rather than directly reproducing the source
trajectories or their scene geometry. Simulation can also expand the data and objectives available for robot
learning. MimicGen~\cite{mimicgen} transforms demonstrations to generate
additional training trajectories, while Re$^3$Sim~\cite{re3sim}
reconstructs real scenes to generate simulation data for imitation
learning. Eureka~\cite{eureka} generates reward code for reinforcement
learning. GenSim~\cite{gensim} and GenSim2~\cite{gensim2} generate
simulation tasks and demonstrations for policy training, with GenSim2
using multimodal reasoning to construct task solvers. RPG uses
simulation to develop and refine executable skills and the system
prompt rather than to train a neural policy. Model weights, task
definitions, and evaluation criteria remain fixed during practice.

\begin{figure*}[t]
\label{sec:transfer-case}
    \centering
    \includegraphics[width=1\textwidth]{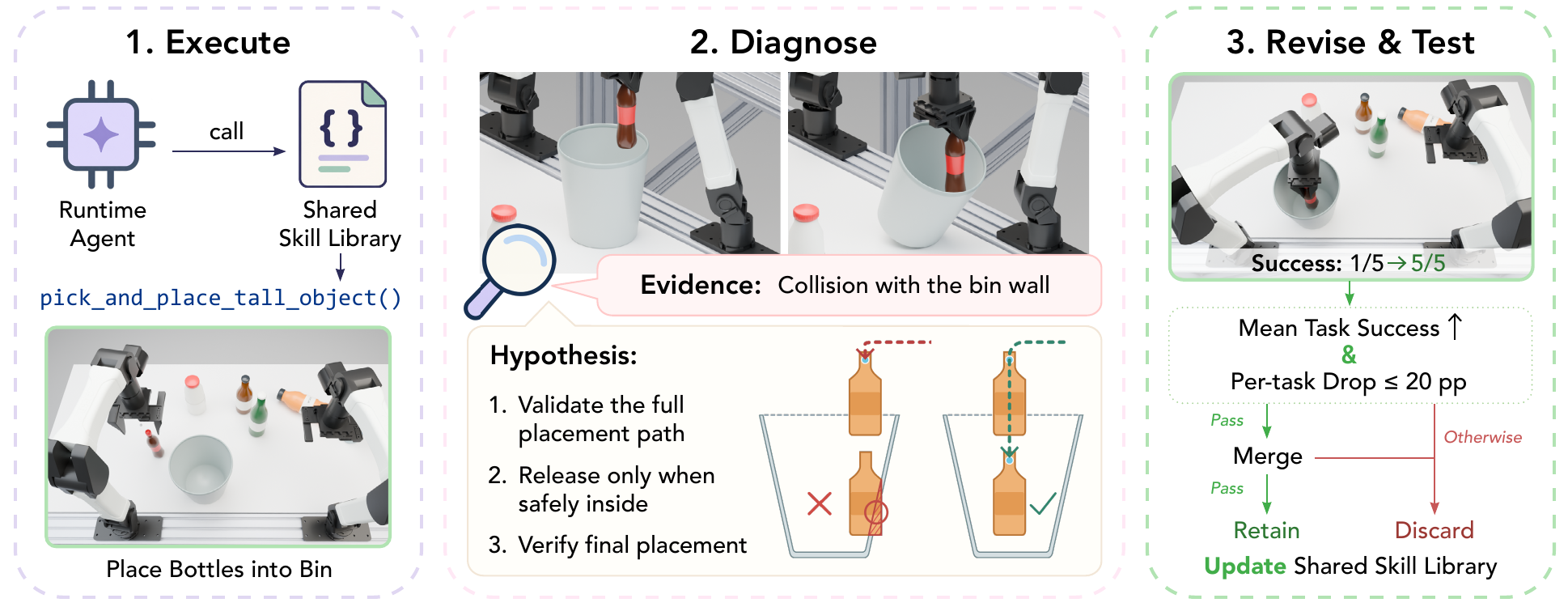}
    \vspace{-1.5em}
    \caption{\textbf{Case study of self-improvement on bottle placement.}
(1) The Runtime Agent attempts the task by calling the shared skill
\texttt{pick\_and\_place\_tall\_object()}, but the rollout fails when the
bottle collides with the bin wall.
(2) Failure diagnosis attributes the error to insufficient placement-path
validation and proposes a more robust procedure: validate the full trajectory,
release only after the bottle is safely inside the bin, and verify the final
placement.
(3) The revised skill improves task success from \textbf{1/5 to 5/5} on the
development trials.
The candidate is eligible for merging only if mean task success increases while no individual task drops by more than 20 percentage points. Selected eligible revisions are then merged, and the merged system is retained only if it passes the same cross-task validation.
}
\label{fig:diagnose-validate}
\vspace{-0.5em}
\end{figure*}

\section{Reconstruct, Practice, Go Real}
\label{sec:method}

\subsection{Agent-as-Policy Setup and Initialization}

RPG coordinates six specialized agents across its three stages: Constructor, Runtime Agent, Privileged Agent, Video Analyzer, Implementor, and Merger (Fig.~\ref{fig:architecture}). It first constructs simulation practice tasks from an offline dataset, improves a shared robot execution system by practicing these tasks, and ultimately deploys the resulting system on a physical robot. We formulate the execution system as \emph{Agent-as-Policy}: a multimodal agent (the Runtime Agent) follows a system prompt to complete each task by selecting and composing reusable skills from a shared skill library. The skills provide generalizable primitives for perception and robot control, and are constructed and refined through the Reconstruct and Practice stages described below.

Let $\mathcal D$ denote the offline dataset. It contains source tasks, each represented by trajectories with videos and textual descriptions. In our experiments, $\mathcal D$ is instantiated with ABC~\cite{abc}. Let $\mathcal W=\{\mathcal T_i\}_{i=1}^{K}$ denote the collection of practice tasks. Each practice task specifies a simulation scene, a goal, an initial-state distribution, and a binary evaluator $e_i(\tau)$ that checks completion from an episode trajectory $\tau$. Task specifications remain fixed during practice, while object configurations vary between episodes according to the specified distribution. The execution system used in round $r$ is
$S_r=(\mathcal L_r,\rho_r)$,
where $\mathcal L_r$ is the shared skill library and $\rho_r$ is the system prompt that instructs the agent how to use these skills to complete tasks. A skill takes arguments such as the selected arm and target position, executes an action sequence, and returns its exit status and measured effects.
The execution system remains fixed within each episode, while the underlying model weights and agent harness remain unchanged throughout Practice.

We initialize a base library \(\mathcal L_{\mathrm{base}}\) with nine low-level tools from CaP-X~\cite{capx} for state queries, perception, inverse kinematics, motion, and gripper control. During Reconstruct, the Constructor examines trajectories from \(\mathcal D\) and builds six higher-level manipulation skills from these tools: pick, place, pick-and-place, table-supported transfer between arms, move-above, and push. The resulting initial practice library \(\mathcal L_0\) contains 15 tools and skills. Together with the initial system prompt \(\rho_0\), this defines \(S_0=(\mathcal L_0,\rho_0)\), from which all diagnostic ablations begin.

\subsection{Reconstruct: Build and Freeze the Practice Tasks}

The Constructor first selects source tasks from \(\mathcal D\) to cover useful and diverse manipulation capabilities. For each selected source task, it chooses 3–8 trajectories and reviews their videos and textual descriptions. It identifies the manipulation capabilities represented by the source task and proposes a simulation task for practicing them. We give the agent a 7.5-minute budget per source task.

The Constructor builds a MuJoCo~\cite{mujoco} simulation environment with the YAM robot model, using available assets and searching online for additional assets as needed. The practice task may cover part of the source task or an analogous manipulation, rather than reproduce the source task or its exact scene geometry. For example, organizing sunglasses becomes closing a case that starts open and loaded (Table~\ref{tab:construction}). Tasks without a corresponding source task in $\mathcal D$ use available task specifications.

The Constructor writes a \texttt{check\_success} function for each task to implement $e_i(\tau)$. Humans check whether it measures the proposed objective before practice begins. Because the offline dataset contains many more source tasks than those instantiated in a single practice run, Reconstruct can continue expanding \(\mathcal W\) by selecting and processing additional source tasks. We fix the resulting task collection before Practice begins so that task definitions and evaluators do not change during self-improvement.

\begin{table}[!t]
\centering
\caption{Representative correspondences between source tasks in the offline dataset and constructed practice tasks.}
\vspace{-1mm}
\label{tab:construction}
\small
\setlength{\tabcolsep}{4pt}
\renewcommand{\arraystretch}{1.12}
\begin{tabularx}{\columnwidth}{@{}>{\raggedright\arraybackslash}p{1.03in}>{\raggedright\arraybackslash}X@{}}
\toprule
\textbf{Source task} & \textbf{Constructed practice task}\\
\midrule
Organize sunglasses & Close the case; opening and loading are encoded in the initial state.\\
Organize makeup & Close the drawer; preceding manipulation is already satisfied.\\
Roll towels & Fold once; a cloth manipulation analogue with a different final shape.\\
No corresponding source task & Transfer an object between arms using a task specification.\\
\bottomrule
\end{tabularx}
\vspace{-3mm}
\end{table}

\subsection{Practice: Improve the System from Execution}

\parahead{Execute tasks.}
At the start of round $r$, the Runtime Agent attempts to solve each task five times with $S_r$. It selects skills and arguments based only on observations available on the deployed robot and its execution history. In parallel, the Privileged Agent runs separate episodes from the same initial states using the same model and skill library. The only difference is that the Privileged Agent also receives simulator state for selecting skills and arguments: object names, poses, sizes, velocities, and cloth-corner positions. Both agents invoke the same skills, which use only sensing and control available on the deployed robot.

Asymmetric actor-critic learning uses full simulator state to train a critic while restricting the actor to images~\cite{asymmetric}. RPG instead uses privileged state in a separate execution to support failure diagnosis. With privileged state information, the Privileged Agent may complete some tasks more reliably than the Runtime Agent using the current skills. Its successful executions can guide improvements to the Runtime Agent, while failures shared by both agents can expose defects in the shared skills.

\parahead{Analyze failures.}
On development tasks, which provide task-specific improvement feedback, the Video Analyzer compares failed Runtime Agent executions with Privileged Agent executions and related source videos from $\mathcal D$ when available. It receives sampled frames, skill calls and return values, perception results, commanded and measured robot states, simulator ground truth, and evaluator metrics. It uses skill calls, gripper events, motion, and annotations from $\mathcal D$ to identify manipulation phases, then compares phases involving the same object role, arm, and operation.

The Video Analyzer identifies the first observed failure and recommends a concrete change, supporting its diagnosis with frames or traces and describing the expected physical effect of the change. It does not modify the execution system. Its diagnosis and recommendation are passed to the Implementor for revision.

\parahead{Revise the system.}
The Implementor receives $S_r$, the execution records, and the Video Analyzer's recommendations. It proposes new skills, changes to existing skills, and revisions to the system prompt. For each development task with fewer than five successful Runtime Agent trials, the Implementor can produce at most one candidate revision. All candidates branch independently from $S_r$. Before testing, we check that each candidate loads, has valid syntax and interfaces, and uses only observations available at deployment. Validation and feedback-held-out tasks do not generate task-specific candidate revisions, but remain part of cross-task validation as regression checks.

\subsection{Validate and Retain System Revisions}

A shared-system revision can help one task but hurt another. We therefore evaluate each candidate using complete Runtime Agent episodes across $\mathcal W$, then merge selected revisions and evaluate the resulting system again.

\parahead{Test candidates.}
We compare each candidate $\widetilde S$ with $S_r$ using the Runtime Agent on the same five fixed development seeds per task. In the reported run, this gives $22\times5=110$ episodes per system. A candidate passes if mean task success increases and no task loses more than one success out of five. For $n=5$, the observed success rate on task $i$ and the mean across tasks are
\begin{equation}
\widehat q_i(S)=\frac1n\sum_{k=1}^{n}e_i(\tau_{ik}(S)),\qquad
\widehat J(S)=\frac1K\sum_{i=1}^{K}\widehat q_i(S).
\label{eq:empirical}
\end{equation}
Here $\tau_{ik}(S)$ is the $k$-th Runtime Agent trajectory on task $i$ with execution system $S$. Let $\Delta\widehat J=\widehat J(\widetilde S)-\widehat J(S_r)$ and $\Delta\widehat q_i=\widehat q_i(\widetilde S)-\widehat q_i(S_r)$. The test is
\begin{equation}
\mathsf{Gate}(\widetilde S;S_r)=\mathbf1\!\left[
\Delta\widehat J>0\ \land\ \min_i\Delta\widehat q_i\geq-0.20
\right].
\label{eq:gate}
\end{equation}
A candidate that passes is eligible for merging, but may still be rejected later based on the execution records.

\parahead{Merge and retest.}
The Merger selects eligible revisions and merges them into $S^+$. Because edits can conflict, we evaluate $S^+$ with the Runtime Agent and apply the same test against $S_r$. If the merged system passes and the Merger chooses to retain it, $S_{r+1}=S^+$. Otherwise, $S_{r+1}=S_r$. Figure~\ref{fig:diagnose-validate} illustrates this procedure on
Place Bottles in Bin, from failure diagnosis and skill revision through cross-task validation, merging, and retention.

We run 15 practice rounds with autonomous revision, validation, merging, and retention. Humans handle setup, evaluator checks, the decision to stop practice, and hardware adaptation. After practice ends, we evaluate the saved round-end systems on separate held-out seeds (Sec.~\ref{sec:experiments}).

\subsection{Go Real: Adapt and Freeze for Deployment}

The final RPG system and CaP-Agent0 are adapted to the physical YAM robot before evaluation. Both receive the same coordinate calibration and hardware adaptation, including pad offsets and the empty-gripper aperture. We do not apply any object- or task-specific tuning. We use \texttt{low} reasoning effort for all physical evaluations to reduce inference cost and latency. Beyond these adaptations, we make no behavioral code changes based on physical feedback. The adapted systems remain fixed during evaluation. The Privileged Agent and Video Analyzer are not used during physical deployment.

\begin{figure*}[!t]
\centering
\includegraphics[width=1\textwidth]{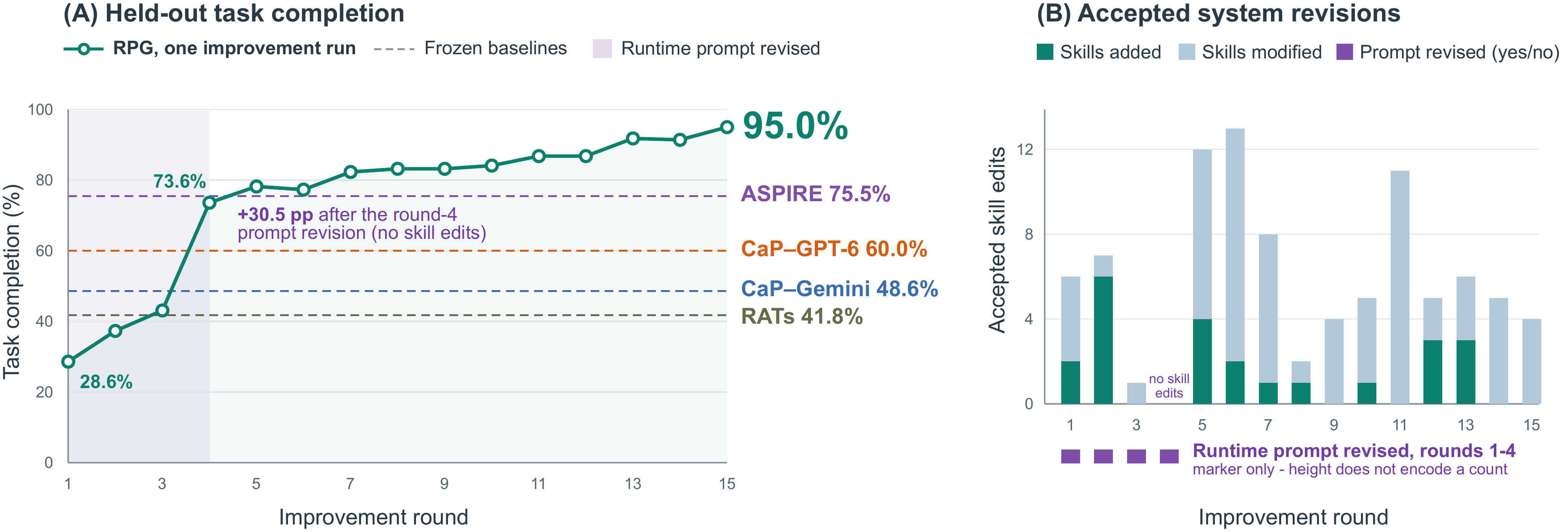} \vspace{-15pt}
\caption{\textbf{Task completion and revision activity.} (A) Round-end systems are evaluated retrospectively on the same 22 tasks $\times$ 10 held-out seeds after development is frozen. Dashed lines show ASPIRE, RATs, and the matched-model and strongest CaP-Agent0 configurations (abbreviated CaP). Table~\ref{tab:per-task} gives the full model comparison. (B) Stacked bars count accepted skill additions (green) and modifications (light blue) in each round. Purple squares mark rounds in which the Runtime Agent's system prompt was revised (rounds 1--4); their height does not encode a count. The curve represents one improvement run. Rounds do not represent matched computational budgets.}
\label{fig:improvement-curve}
\label{fig:revision-activity}
\vspace{-1em}
\end{figure*}

\section{Evaluation}
\label{sec:experiments}
\label{sec:results}
We organize the evaluation around four questions: \textbf{Q1}, how does task success change with practice, and how does RPG compare with existing agents? \textbf{Q2}, how do the diagnostic inputs affect improvement? \textbf{Q3}, can library changes improve task completion with the surrounding runtime fixed? \textbf{Q4}, does the resulting system complete physical tasks after common adaptation?

\ssecv
\subsection{Tasks and Evaluation Protocol}
\ssecv
The simulated YAM robot has two 6-DoF arms and two grippers, and receives RGB-D observations. It accepts 14-dimensional commands specifying absolute joint-position targets. The practice collection contains seven CaP-X tasks and fifteen ABC-derived or analogous tasks, covering transport, sorting, stacking, extraction, insertion, articulated closing, bimanual manipulation, and cloth folding.

Task specifications identify which bodies are randomized. Default planar displacement and yaw bounds are $\pm3$~cm and $\pm0.5$~rad for rigid bodies and $\pm4$~cm and $\pm0.3$~rad for cloth. Perturbations also include table height ($\pm8$~mm), camera translation ($\pm4$~mm), camera rotation ($\pm0.008$~rad), and depth noise and bias. Case and drawer fixtures stay fixed. Development and evaluation sample the same fixed task distributions, with paired initializations across methods.

\parahead{Separate development and evaluation.} Practice uses five fixed development seeds per task for candidate and integration checks. After the 15-round practice run is complete and its saved systems are frozen, we retrospectively evaluate each round-end system on the same ten held-out seeds per task, giving 220 episodes per point in Fig.~\ref{fig:improvement-curve}. These seeds are excluded from improvement and checkpoint selection. The frozen baselines in Table~\ref{tab:per-task} use the same evaluation initializations. Because each task has the same number of evaluation episodes, mean task success rate equals pooled episode success rate.

\parahead{Success criteria and evaluation budget.}
Each task is evaluated using its predefined success condition. Most conditions directly check whether the target object reaches its specified goal state. Bottle placement succeeds only when both target bottles are placed correctly among the distractors. Towel folding succeeds when the final footprint is 42--58\% of its initial area, corresponding corners are within 5~cm, the moved layer is at least 3~mm above the target corners, and the maximum thickness is below 5~cm. Each episode has a 20-minute robot-time limit, in addition to the model-interaction and wall-time limits specified below.

\begin{table*}[!t]
\centering
\vspace{2mm}
\caption{Simulation task completion on ten held-out seeds per task.
CaP-Agent0 and ASPIRE are reproduced from their released code.
The RATs column reports any-time success.
Bold marks the highest observed count in each row.}
\label{tab:per-task}
\vspace{-1mm}

\begingroup
\small
\setlength{\tabcolsep}{5pt}
\renewcommand{\arraystretch}{1.15}

\setlength{\aboverulesep}{2pt}
\setlength{\belowrulesep}{2pt}
\setlength{\cmidrulekern}{4pt}
\arrayrulecolor{black}

\newcommand{\modelcell}[1]{%
  \parbox[c][6ex][c]{\linewidth}{\centering #1}%
}

\newsavebox{\rpgtablebox}
\newdimen\rpgshadewidth

\rpgshadewidth=\dimexpr
  (\textwidth-1.75in-14\tabcolsep)/7+\tabcolsep
\relax

\sbox{\rpgtablebox}{%
\begin{tabularx}{\textwidth}{
  @{}p{1.75in}
  *{7}{>{\centering\arraybackslash}X}
  @{}
}
\toprule

\rule{0pt}{5.8ex}
& \multicolumn{4}{c}{
    \shortstack{
      \textbf{CaP-Agent0}\\[1pt]
      {50 turns}
    }
  }
& \shortstack{
    \textbf{RATs}\\[1pt]
    {50 turns}
  }
& \shortstack{
    \textbf{ASPIRE}\\[1pt]
    {30 turns}
  }
& \shortstack{
    \textbf{RPG}\\[1pt]
    {30 turns}
  } \\[-1pt]

\cmidrule(lr){2-5}
\cmidrule(lr){6-6}
\cmidrule(lr){7-7}
\cmidrule(l){8-8}

\parbox[c][6ex][c]{\linewidth}{\textbf{Task}}
& \modelcell{Gemini 3.8\\Flash}
& \modelcell{GPT-6\\Astra Pro}
& \modelcell{Opus 5}
& \modelcell{Fable 5.1}
& \modelcell{Gemini 3.8\\Flash}
& \modelcell{Gemini 3.8\\Flash}
& \modelcell{Gemini 3.8\\Flash} \\
\midrule

\inputtable tables/per_task_current.tex

\bottomrule
\end{tabularx}%
}

\resizebox{1.0\linewidth}{!}{%
  \makebox[0pt][l]{%
    \hspace*{\dimexpr
      \wd\rpgtablebox-\rpgshadewidth
    \relax}%
    {\color{paperbluefill}%
      \rule[-\dp\rpgtablebox]
        {\rpgshadewidth}
        {\dimexpr\ht\rpgtablebox+\dp\rpgtablebox\relax}%
    }%
  }%
  \usebox{\rpgtablebox}%
}

\endgroup
\vspace{-2mm}
\end{table*}

\ssecv
\subsection{Q1: Does RPG Improve with Practice?}
\ssecv
\label{sec:simulation-results}
\parahead{Baselines.} We reproduce CaP-Agent0 from CaP-X~\cite{capx} and ASPIRE~\cite{aspire} using their released code, and evaluate RATs~\cite{playful} after 50 iterations of self-directed play. CaP-Agent0 is evaluated with the four models listed in Table~\ref{tab:per-task}, while RPG, ASPIRE, and RATs use Gemini 3.8 Flash for online decision making and task-level code generation. All models use \texttt{xhigh} reasoning effort in the simulation evaluations.

\parahead{Evaluation budgets.} CaP-Agent0 receives up to 50 model turns per episode. RATs receives up to 50 model turns and a 1,200~s wall-clock time limit. RPG and ASPIRE receive up to 30 model turns, 12 model-based observation calls, and 1,200~s of wall time.

\parahead{Self-improvement models.}
RPG uses Fable 5.1 for offline code revision during self-improvement, including execution-video analysis and updates to the shared execution system between episodes, while Gemini 3.8 Flash generates task-level programs during online execution.

\parahead{Task completion.} As shown in Fig.~\ref{fig:improvement-curve} and Table ~\ref{tab:per-task}, RPG achieves 209/220 successes (95.0\%), outperforming all baselines. Among the four models evaluated with CaP-Agent0, GPT-6 Astra Pro performs best with 132/220 successes (60.0\%), below RPG's performance with Gemini 3.8 Flash. Gemini CaP-Agent0 achieves 107/220 (48.6\%), ASPIRE achieves 166/220 (75.5\%), and RATs achieves 92/220 (41.8\%). 

Compared with Gemini CaP-Agent0, RPG improves on 19 tasks and ties on the remaining three. The improvement spans both task groups: success increases from 51.4\% to 98.6\% on CaP-X tasks and from 47.3\% to 93.3\% on ABC-derived or analogous tasks.

RPG succeeds in all ten trials on 17 of the 22 tasks. Compared with ASPIRE, its largest gains come from fixture insertion and bottle placement, which together account for 15 of its 43 additional successes. ASPIRE succeeds in one more trial than RPG on serving and standard cube sorting. Seven of RPG's 11 failures occur in towel folding, making deformable manipulation the main remaining failure mode in this evaluation.

\parahead{Improvement across rounds.} RPG increases from 63/220 successes after round 1 (28.6\%) to 209/220 after round 15 (95.0\%). Round 4 exceeds all four CaP-Agent0 configurations and round 5 reaches 78.2\%, exceeding ASPIRE's 75.5\% endpoint (Fig.~\ref{fig:improvement-curve}A).

\parahead{What changes in the system?} Over 15 practice rounds, the skill library grows from 15 to 38 entries, with 23 new skills and 66 modifications to existing skills (Fig.~\ref{fig:revision-activity}B). The system prompt is revised during the first four rounds. The largest improvement occurs between rounds 3 and 4, when success increases from 43.2\% to 73.6\%. This round changes the system prompt but records no skill edit. The new prompt introduces a bounded perception--action loop and tracks the remaining interaction budget, encouraging the agent to act on its current estimates instead of repeatedly requesting observations. Because the same prompt revision also changes how the agent uses its observation budget, the 30.5-percentage-point gain reflects the combined behavioral effect of the prompt update rather than any single instruction.

After the system prompt is fixed, success rises by another 21.4 percentage points between rounds 4 and 15. During this period, the system makes 60 modifications to existing skills. In rounds 9, 11, 14, and 15, it modifies existing skills without adding new ones. RPG therefore improves both by expanding the library and by repairing shared skill implementations.

The evaluation curve is not strictly monotonic: the number of successes decreases by one at rounds 6 and 14. These revisions passed the five-seed development checks but did not improve every result on the ten held-out evaluation seeds.

\ssecv
\subsection{Q2: Which Diagnostic Inputs Drive Improvement?}
\ssecv
\label{sec:guidance-ablation}
We compare four variants: full RPG, RPG without the Video Analyzer, RPG without the Privileged Agent, and RPG without either component. All variants start from the same system, consisting of nine base tools, six skills constructed from trajectories in the offline dataset, and the same runtime configuration. Each variant undergoes five improvement rounds using the same models, initial prompts, per-round budgets, and candidate and integration gates. We then evaluate the frozen variants on the same ten held-out seeds per task.

Removing the Video Analyzer disables its failure analysis and its access to videos from the offline dataset during practice, while Runtime Agent traces and available Privileged Agent feedback remain. Removing the Privileged Agent disables its rollouts and feedback. In all variants, the six skills constructed from these trajectories remain part of the initial library.

\begin{table}[t]
\centering
\caption{Five-round diagnostic ablations from the same 15-skill initialization. Each variant is evaluated on 220 held-out episodes.}
\label{tab:guidance-ablation}
\vspace{-1mm}
\begingroup
\small
\setlength{\tabcolsep}{4pt}
\renewcommand{\arraystretch}{1.18}
\begin{tabularx}{\columnwidth}{
  >{\hsize=1.14\hsize\linewidth=\hsize
    \centering\arraybackslash}X
  >{\hsize=1.14\hsize\linewidth=\hsize
    \centering\arraybackslash}X
  >{\hsize=.72\hsize\linewidth=\hsize
    \centering\arraybackslash}X
}
\toprule
\textbf{Video Analyzer}
& \textbf{Privileged Agent}
& \textbf{Success}\\
\midrule
\rowcolor{paperbluefill}
Yes & Yes & \textbf{78.2\%}\\
No  & Yes & 51.8\%\\
Yes & No  & 50.9\%\\
No  & No  & 50.9\%\\
\bottomrule
\end{tabularx}
\endgroup
\vspace{-3mm}
\end{table}

Full RPG achieves 172/220 successes (78.2\%), compared with 114/220 (51.8\%) without the Video Analyzer and 112/220 (50.9\%) without the Privileged Agent or without both components. Removing the Video Analyzer reduces success by 26.4 percentage points, while removing the Privileged Agent reduces it by 27.3 points. Each single-component variant performs within 0.9 percentage points of the variant with neither component, suggesting that the two provide complementary information. The Privileged Agent supplies reference executions using privileged state, while the Video Analyzer uses them together with Runtime Agent traces and videos from the offline dataset to diagnose failures.

\begin{figure*}[!t]
\centering
\includegraphics[width=0.92\textwidth]{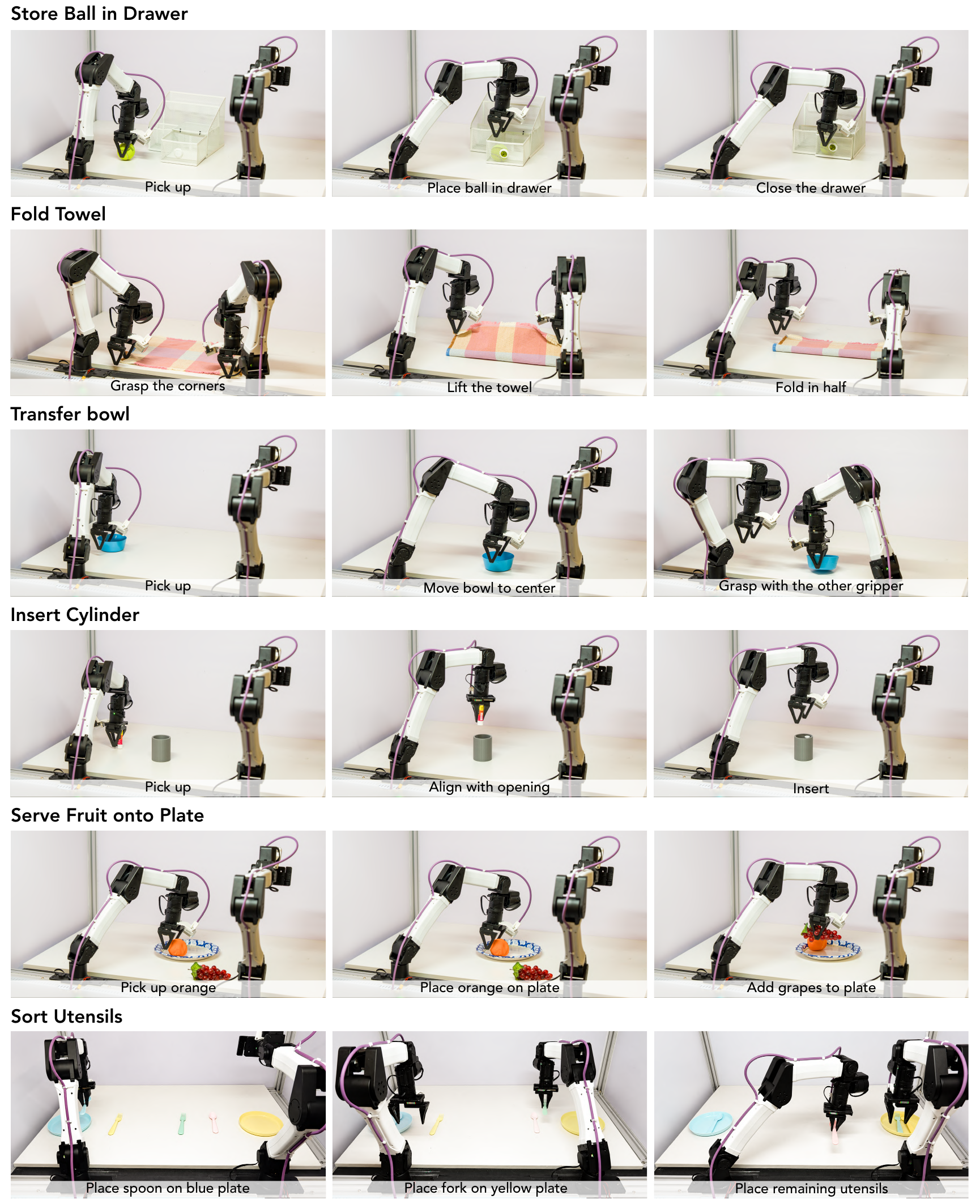}
\vspace{-2mm}
\caption{\textbf{Physical task execution.}
Representative execution sequences on the physical YAM robot.
The first three rows show the quantitatively evaluated tasks: Store Ball in Drawer, Fold Towel, and Transfer Bowl. Bowl transfer uses table-supported staging.
The remaining rows show additional qualitative deployments on Insert Cylinder, Serve Fruit onto Plate, and Sort Utensils.
Table~\ref{tab:real} reports outcomes for the three quantitative tasks.}
\label{fig:real-sequences}
\vspace{-5mm}
\end{figure*}

\ssecv
\subsection{Q3: Do Skill-Library Revisions Improve Task Success?}
\ssecv
\label{sec:skill-setup}
\label{sec:intervention-results}

The main improvement trajectory updates multiple parts of
the execution system. To isolate executable skill improvement
from changes in prompting or online reasoning, we compare two
recorded library snapshots while fixing the runtime model
(Gemini 3.8 Flash with xhigh reasoning), prompt, perception
implementation, and episode budgets. We evaluate both libraries on the same ten initializations of four tasks. This controlled comparison is separate from the final 22-task benchmark and uses library snapshots different from those involved in the Round-12 development transition illustrated in Fig.~\ref{fig:diagnose-validate}.

The revised library modifies skills for coordinated motion, lifting, and pick clearance, and adds new skills for rim grasping, supported set-down, and peg insertion.
Three tasks directly test these revisions, while bottle placement checks whether they preserve the existing tall-object behavior. As shown in Table~\ref{tab:intervention}, success improves on all three targeted tasks and remains at 10/10 for bottle placement. Overall success increases from 26/40 to 37/40, a gain of 27.5 percentage points.

\begin{table}[tb]
\centering
\caption{Library-only comparison before and after the library revisions, with ten paired initializations per task. The runtime remains fixed, and only the library changes.}
\label{tab:intervention}
\vspace{-1mm}
\begingroup
\small
\setlength{\tabcolsep}{4pt}
\renewcommand{\arraystretch}{1.18}
\begin{tabularx}{\columnwidth}{
  @{}
  >{\hsize=1.38\hsize\linewidth=\hsize
    \raggedright\arraybackslash}X
  >{\hsize=.81\hsize\linewidth=\hsize
    \centering\arraybackslash}X
  >{\hsize=.81\hsize\linewidth=\hsize
    \centering\arraybackslash}X
  @{}
}
\toprule
\textbf{Task}
& \textbf{Before repair}
& \textbf{After repair}\\
\midrule
Two-arm lift
& 6/10 & \textbf{10/10}\\
Place dish on rack
& 5/10 & \textbf{9/10}\\
Nut assembly
& 5/10 & \textbf{8/10}\\
Place bottles in bin
& \textbf{10/10} & \textbf{10/10}\\
\midrule
\textbf{Total}
& {26/40} & \textbf{37/40}\\
\textbf{Success rate}
& {65.0\%} & \textbf{92.5\%}\\
\bottomrule
\end{tabularx}
\endgroup
\vspace{-4mm}
\end{table}

\parahead{Why do the library revisions help?}
To examine one repair in the controlled library-only comparison, we inspect
the lifting behavior used by several tasks.
Before revision, the lift skill requests a 6~cm displacement but moves only
\mbox{4.6--5.1~cm}, while reporting the commanded displacement as the
achieved displacement.
It also defines the lift target relative to a nominal grasp pose, even when
the preceding approach terminates early.
The revised skill instead completes the approach, defines the lift relative
to the measured post-grasp fingertip position, and verifies the actual
displacement, grasp retention, and object levelness.
Among lift-skill invocations observed in these controlled episodes, 12 of 13 fail before the revision, whereas all 11 observed invocations succeed afterward.
The task-level results in Table~\ref{tab:intervention} show how this repair,
together with the other library revisions, affects complete-task performance.

We further test both libraries with the same privileged scripted caller,
which is separate from the LLM-based Privileged Agent used during Practice.
Changing only the library increases success from 19/40 to 39/40, providing
additional evidence that the executable revisions themselves improve task
completion.
The gains are not uniform, however.
Both libraries pass 12 of 14 auxiliary runtime checks, with the revised
library gaining one success on one task but losing one on ABC sorting.
Adding five paired trials yields pooled ABC-sorting results of 6/6 before
the revision and 5/6 afterward.
This local regression reinforces the need to evaluate shared-skill revisions
through their effect on complete task-level behavior.

\subsection{Q4: Does RPG Transfer to a Physical Robot?}

\label{sec:physical-results}

\parahead{Protocol.} We evaluate three physical tasks on a YAM robot: placing a ball in a drawer and closing it, folding a towel, and transferring a bowl between hands (Fig.~\ref{fig:real-sequences}). Each method receives ten trials per task. We compare the final RPG system, using Gemini 3.8 Flash, with CaP-Agent0 using either Gemini 3.8 Flash or GPT-6 Astra Pro. All methods use the same calibration and hardware-adaptation procedure, including a single uniform grasp offset without object- or task-specific tuning. Each trial uses \texttt{low} reasoning effort and terminates after 30 VLM calls or 20 minutes, whichever comes first.

Object positions are randomized, and the same initial arrangements are manually reproduced across methods. The ball begins outside an open drawer, the towel lies unfolded on the table, and the bowl starts on the left side of the workspace. The robot may autonomously retry or recover within a trial, but no human assistance, manual reset, or system revision is allowed. We measure both intermediate-stage and complete-task success within the same trials.

\parahead{Success criteria.} For ball placement and drawer closure, the intermediate stage succeeds when the ball is placed inside the open drawer. The complete task succeeds when the ball remains inside after the drawer is fully closed. Towel folding succeeds when the estimated final-to-initial footprint ratio is between 0.40 and 0.60 and the folded towel has approximately parallel sides. For bowl transfer, the intermediate stage succeeds when the left hand lifts the bowl and moves it to the center of the workspace. The complete task succeeds when the right hand alone holds the bowl off the table for at least three seconds after the left hand releases it. Intermediate support from the table is allowed. The physical folding criterion is defined independently of the simulation criterion and is not intended to match its difficulty exactly.

\begin{table}[t]
\centering
\caption{Physical outcomes after a common calibration and hardware-adaptation procedure.
Each method receives ten trials per task. Intermediate-stage and complete-task
outcomes are measured in the same trials.}
\label{tab:real}
\vspace{-1mm}

\begingroup
\small
\setlength{\tabcolsep}{4pt}
\renewcommand{\arraystretch}{1.16}
\setlength{\aboverulesep}{0pt}
\setlength{\belowrulesep}{0pt}

\begin{tabularx}{\columnwidth}{
  @{}
  >{\hsize=1.84\hsize\linewidth=\hsize
    \raggedright\arraybackslash}X
  >{\hsize=.72\hsize\linewidth=\hsize
    \centering\arraybackslash}X
  >{\hsize=.72\hsize\linewidth=\hsize
    \centering\arraybackslash}X
  >{\hsize=.72\hsize\linewidth=\hsize
    \centering\arraybackslash
    \columncolor{paperbluefill}[\tabcolsep][0pt]}X
  @{}
}
\toprule
& \multicolumn{2}{c}{\textbf{CaP-Agent0}}
& \textbf{RPG}\\
\textbf{Task / Stage}
& Gemini & GPT-6 & Gemini\\
\midrule

\rule{0pt}{2.6ex}\textbf{Store ball in drawer}
& 3/10 & 4/10 & \textbf{10/10}\\
\hspace*{0.6em}Place ball in drawer
& 7/10 & 6/10 & \textbf{10/10}\\

\rule{0pt}{3ex}\textbf{Fold towel}
& 0/10 & 9/10 & \textbf{10/10}\\

\rule{0pt}{3ex}\textbf{Transfer bowl}
& 0/10 & 1/10 & \textbf{10/10}\\
\hspace*{0.6em}Move bowl to center
& 6/10 & 8/10 & \textbf{10/10}\\

\bottomrule
\end{tabularx}
\endgroup

\vspace{-4mm}
\end{table}

\begin{figure*}[t]
    \centering
    \includegraphics[width=0.92\textwidth]{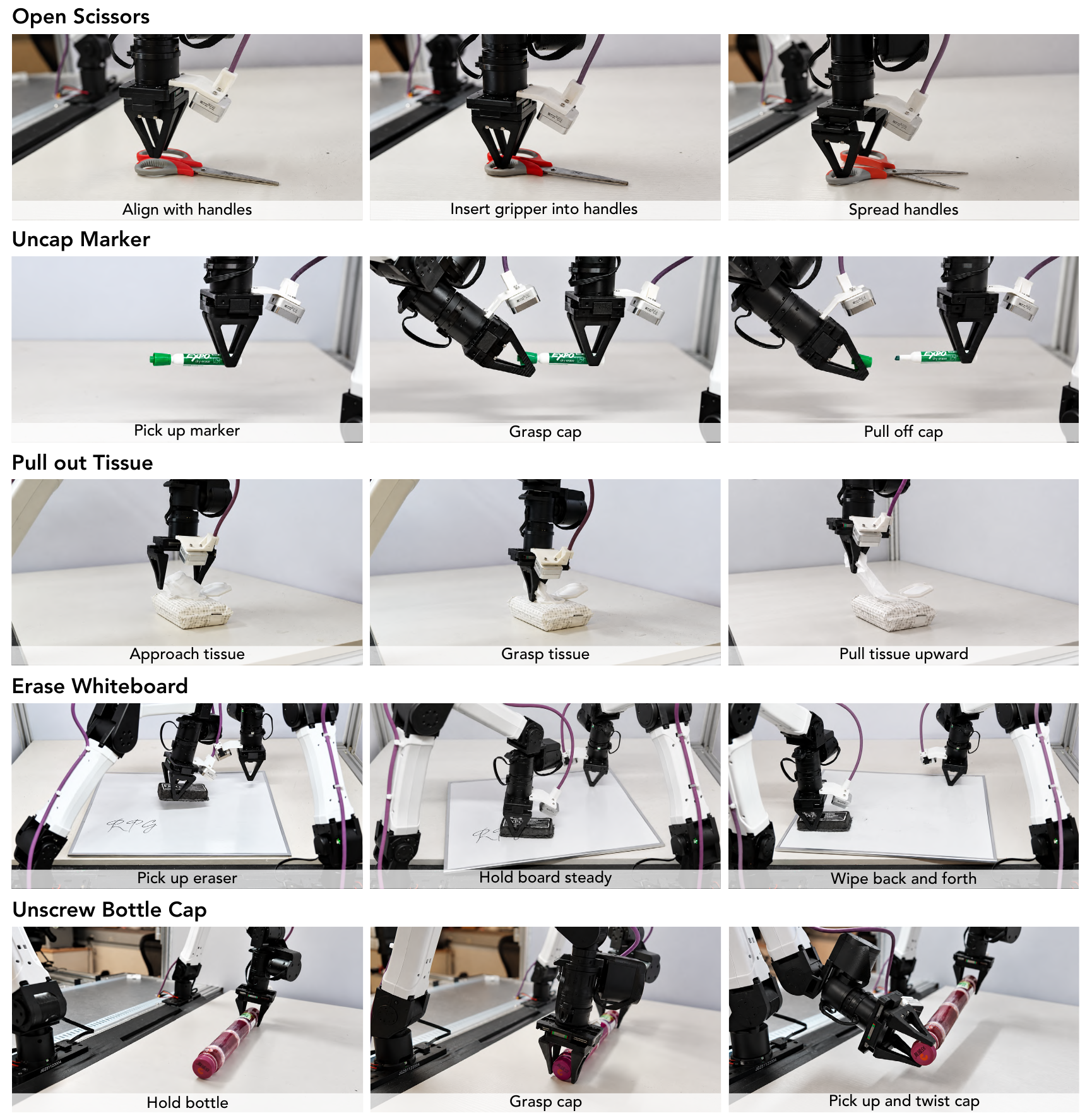}
    \vspace{-2mm}
    \caption{\textbf{Qualitative zero-shot real-world deployment.}
    Representative execution sequences for five additional physical tasks:
    Open Scissors, Uncap Marker, Pull out Tissue, Erase Whiteboard, and
    Unscrew Bottle Cap.
    The frozen RPG system receives the task instruction and executes each task
    without task-specific real-world tuning, demonstrations, or retry-based
    adaptation.
    These examples are qualitative and are not included in the quantitative
    baseline comparison.}
    \label{fig:zero_shot_real}
\vspace{-4mm}
\end{figure*}

\parahead{Analysis.} RPG completes all ten trials of each physical task (Table~\ref{tab:real}). Among the CaP-Agent0 variants, GPT-6 Astra Pro shows stronger multi-stage task completion than Gemini 3.8 Flash. Gemini places the ball inside the drawer in 7/10 trials but completes drawer closure in only 3/10, and it moves the bowl to the transfer position in 6/10 trials without completing any handover. GPT-6 Astra Pro improves complete-task success to 4/10 for the drawer task and 1/10 for bowl transfer. It also folds the towel in 9/10 trials.

These results suggest that improving the shared execution system can provide larger gains than replacing the runtime model alone. RPG's advantage is largest on the two multi-stage tasks. On towel folding, it exceeds GPT-6 Astra Pro CaP-Agent0 by only one success. All reported outcomes include any autonomous retries or recovery within the stated episode budget.

\paragraph{Zero-shot real-world deployment.}
Beyond the three quantitatively evaluated tasks, we qualitatively deploy the
frozen RPG system on five additional physical tasks:
Open Scissors, Uncap Marker, Pull out Tissue, Erase Whiteboard, and Unscrew Bottle Cap
(Fig.~\ref{fig:zero_shot_real}).
These deployments use no task-specific real-world tuning, demonstrations,
or retry-based adaptation.
The system receives only the task instruction at deployment and executes it
by composing the shared skills developed during Practice.
We report these examples qualitatively and do not use them for quantitative
comparison with the baselines.
Together, they illustrate that the frozen execution system can be applied to
new physical task instructions without task-specific real-world adaptation.

\section{Conclusion}

RPG diagnoses failures using privileged execution and validates revisions across tasks to improve a persistent robot execution system without updating model weights. Success rises from 28.6\% after the first practice round to 95.0\% after 15 rounds on held-out initializations of 22 tasks, outperforming ASPIRE (75.5\%) and CaP-Agent0 with GPT-6 Astra Pro (60.0\%). Ablations support combining the Video Analyzer and Privileged Agent, while a library-only comparison isolates gains from skill revisions and highlights the importance of task-level validation. Following the same calibration and hardware-adaptation procedure as the baselines, the frozen system succeeds in all 30 trials across three physical tasks. These results show that shared skill code and the system prompt can serve as practical targets for persistent robot improvement.

\noindent\textbf{Discussion and Limitations.} Two bottlenecks limit self-improvement. Each round collects Runtime Agent and Privileged Agent trajectories across all 22 tasks, but API rate limits constrain parallel rollouts. As the task set grows, overlapping revisions to shared skills and dependencies can also complicate integration. Future work could improve efficiency through quota-aware rollout scheduling and dependency-aware proposal integration.

\section{Acknowledgments}
We thank Zi Wang and Bike Zhang for helpful discussions and feedback. OpenAI ChatGPT and Anthropic Claude were used to assist with language editing and manuscript polishing. All technical content, analyses, and conclusions were developed and verified by the authors.

\bibliographystyle{IEEEtran}
\bibliography{references}

\newpage
\clearpage
\appendix

\subsection{Self-Improvement Protocol}
\label{app:self_improvement_protocol}

RPG improves a persistent execution system over successive practice rounds.
We denote the system used in Round~$r$ as
$S_r=(\mathcal L_r,\rho_r)$, where $\mathcal L_r$ is the shared skill library and
$\rho_r$ is the system prompt.
Model weights, task definitions, initial-state distributions, and evaluation
criteria remain fixed throughout Practice.
The execution system is also fixed within an episode and can only be revised
between rounds.

\paragraph{Practice rollouts.}
At the beginning of each round, the Runtime Agent attempts every practice task
on five fixed development initializations.
It selects and composes reusable skills using only observations available at
deployment time and its execution history.
In parallel, a Privileged Agent runs separate episodes from matched initial
states using the same runtime model and the same shared skill library.
The Privileged Agent differs only in its access to simulator state, including
object identities, poses, sizes, velocities, and cloth-corner positions.
Both agents invoke the same deployment-compatible skills. The
skills themselves do not receive privileged state.

\paragraph{Failure diagnosis.}
Failed executions on development tasks are analyzed by a Video Analyzer.
The analyzer compares executions from the Runtime Agent and Privileged Agent and, when available, related source videos from the offline dataset.
Its inputs include sampled frames, skill calls and return values, perception
outputs, commanded and measured robot states, simulator ground truth, and
evaluator metrics.
It identifies the first observable failure, attributes a likely cause, and
proposes a concrete correction supported by execution evidence.
The analyzer does not directly modify the execution system. Its diagnosis is
passed to the Implementor.

\paragraph{Candidate generation.}
The Implementor can revise shared skill implementations and the system
prompt.
A development task can generate a candidate only when the Runtime Agent succeeds on fewer than all five development trials, and at most one candidate is generated for that task in a round. Validation and feedback-held-out tasks do not generate task-specific candidate revisions, but remain in cross-task validation as regression checks.
Development tasks already achieving $5/5$ do not generate new candidates, but remain part of cross-task validation and can therefore detect regressions in shared components.

\paragraph{Cross-task candidate validation.}
Because a revision to a shared skill may improve one task while degrading
others, candidates are evaluated on complete Runtime Agent episodes across the
entire 22-task practice suite.
The candidate and the round-start reference system are evaluated on the same
five fixed development initializations for every task, giving
$22\times5=110$ episodes per system.

Let $\hat{q}_i(S)$ denote empirical success on task $i$ and let
$\hat{J}(S)$ denote mean task success:
\begin{equation}
\hat{q}_i(S)
=
\frac{1}{5}\sum_{k=1}^{5} e_i(\tau_{ik}(S)),
\qquad
\hat{J}(S)
=
\frac{1}{K}\sum_{i=1}^{K}\hat{q}_i(S),
\end{equation}
where $K=22$ and $e_i$ is the predefined binary task evaluator.
For a candidate $\tilde{S}$, define
\begin{equation}
\Delta\hat{J}
=
\hat{J}(\tilde{S})-\hat{J}(S_r),
\qquad
\Delta\hat{q}_i
=
\hat{q}_i(\tilde{S})-\hat{q}_i(S_r).
\end{equation}
A candidate is eligible for integration only if
\begin{equation}
\label{eq:appendix_candidate_gate}
\Delta\hat{J} > 0
\qquad\text{and}\qquad
\min_i \Delta\hat{q}_i \geq -0.20 .
\end{equation}
Thus, a candidate must improve mean task completion without reducing any
task's success count by more than one out of five.

\paragraph{Greedy integration of candidate revisions.}
Multiple candidates may satisfy the cross-task gate in the same round.
Eligible candidates are ranked by their increase in mean development success,
$\Delta\hat{J}$, with deterministic tie breaking.
The highest-ranked candidate initializes the current integrated system.
RPG then considers the remaining eligible candidates in order and greedily
attempts to incorporate them one at a time.
The final implementation permits at most eight such integration attempts in a
round; candidates beyond this budget are not attempted.

Integration is performed only over files modified by the candidate.
A file modified only by the new candidate can be adopted directly.
When both the current integrated system and the new candidate modify the same
file, RPG first performs a three-way merge relative to their common parent.
If the conflict cannot be resolved mechanically, a coding agent resolves the
remaining file-level conflict.
The merger is instructed to preserve compatible additions from both revisions,
including newly introduced reusable skills and their interfaces, while avoiding
unrelated changes, privileged simulator dependencies, or task-specific
hard-coding.
Skill-registration metadata is updated only when the corresponding executable
implementation is present.

\paragraph{Merged-system validation and retention.}
Candidate-level success does not guarantee that individually beneficial revisions remain effective after integration.
The integrated system is therefore evaluated again with complete Runtime Agent
episodes over the same full development suite and is subjected to the same
cross-task regression criterion relative to the accepted round-start system.
If the integrated revision passes the validation and is retained, it becomes
the next round's execution system,
\begin{equation}
S_{r+1}=S^{+}.
\end{equation}
Otherwise, RPG preserves the previous accepted system,
\begin{equation}
S_{r+1}=S_r.
\end{equation}
This rollback preserves the accepted execution system when a candidate or
merged revision fails validation.

\paragraph{One round in summary.}
Each practice round therefore follows the sequence
\begin{enumerate}
    \item execute all practice tasks with the current Runtime Agent and
          Privileged Agent;
    \item diagnose unsuccessful Runtime Agent executions on development tasks;
    \item propose at most one revision for each development task below $5/5$ success;
    \item evaluate each revision across all 22 tasks on five paired
          development initializations;
    \item greedily integrate eligible revisions and retest the combined system;
    \item retain the integrated system only if it passes cross-task validation.
\end{enumerate}
The reported experiment repeats this
rollout--diagnose--revise--validate--integrate procedure for 15 rounds.

\paragraph{Development versus held-out evaluation.}
The five development initializations used for candidate and integration
decisions are separate from the held-out evaluation initializations.
Only after the 15-round practice run is complete and its saved systems are frozen do we
evaluate every saved round-end system retrospectively on ten held-out
initializations per task.
Each point on the reported learning curve therefore contains
$22\times10=220$ held-out evaluation episodes.
These held-out initializations are excluded from candidate generation,
acceptance, integration, and checkpoint selection.
Consequently, the held-out round-by-round curve measures the behavior of the
accepted systems on initializations not used to drive self-improvement.

\paragraph{Models and execution budgets.}
In simulation, both the Runtime Agent and Privileged Agent use
Gemini~3.8~Flash with \texttt{xhigh} reasoning.
Offline failure analysis and code revision use Fable~5.1 with
\texttt{xhigh} reasoning.
Each RPG episode is limited to 30 model turns, 12 model-based observation
calls, 1,200\,s of wall time, and 20 minutes of robot execution time.
The final runtime configuration additionally limits an episode to 400 skill
calls and limits an individual generated program execution to 300\,s.
Model-based runtime queries and visual observation calls share the same 12-call observation budget rather than having separate quotas.

\subsection{Qualitative Examples of Skill Acquisition and Refinement}
\label{sec:appendix_skill_examples}

We present representative examples of how self-improvement changes the
robot's reusable manipulation behaviors. During self-improvement, per-task
development success is measured on five fixed development initializations,
so these in-loop success rates change in increments of 20 percentage points. Candidate
revisions are subsequently evaluated across the full cross-task validation
suite as described in Appendix~\ref{app:self_improvement_protocol}. We
distinguish these in-loop development measurements from the separate
10-seed held-out evaluations reported in the main experiments.

\paragraph{Robust object-in-container placement.}
The Place bottles in bin task provides a clear example of acquiring a
previously unreliable manipulation capability. In \textbf{Round 12}, its
in-loop success rate increased from \textbf{1/5 (20\%) to 5/5 (100\%)}.

Before this improvement, the agent often treated moving the gripper near
the receptacle as sufficient for successful placement. It could commit to
a placement without checking the complete grasp-to-placement trajectory,
continue after an unreachable waypoint, or release a bottle near the bin
without ensuring that the bottle itself had cleared the rim and entered
the opening.

The improved behavior instead localizes the receptacle opening, checks the
complete manipulation route, reasons about the geometry of the bottle
rather than only the fingertip pose, clears the rim before lowering, and
verifies containment after release. In short, the behavior changes from

\begin{quote}
\small
\texttt{pick $\rightarrow$ move near receptacle $\rightarrow$ release}
\end{quote}

to

\begin{quote}
\small
\texttt{localize opening $\rightarrow$ check route $\rightarrow$ verify grasp
$\rightarrow$ align bottle $\rightarrow$ clear rim $\rightarrow$ lower
$\rightarrow$ release $\rightarrow$ verify containment}.
\end{quote}

This represents a transition from end-effector-centric manipulation to
object-centric manipulation: reaching the commanded gripper pose is no
longer treated as equivalent to achieving the desired physical state.

\paragraph{Geometry-aware precision insertion.}
The Nut assembly task illustrates how self-improvement can
turn generic object manipulation into a more precise insertion procedure.
In the five-seed development evaluation, success increased from
\textbf{1/5 (20\%) in Round 4 to 5/5 (100\%) in Round 5}. The system
prompt remained fixed across this transition. The change came from the
reusable manipulation behavior available to the agent.

Before the improvement, the robot could proceed without confirming that
the nut was actually retained, confuse the graspable handle with nearby
geometry, or align the gripper with the peg without accounting for the
location of the nut's square hole. It could also release after an
unsuccessful insertion attempt without verifying seating.

The improved behavior first verifies the grasp with a short test lift,
estimates the offset between the gripper and the square hole, aligns the
\emph{hole} rather than the fingertip with the peg, and inserts through
small monitored vertical increments. If insertion stalls, the robot
retreats and re-observes before continuing. Release occurs only after
visible seating.

\begin{quote}
\small
\texttt{grasp handle $\rightarrow$ test lift $\rightarrow$ verify retention
$\rightarrow$ estimate hole offset $\rightarrow$ align hole to peg
$\rightarrow$ incremental insertion $\rightarrow$ verify seating
$\rightarrow$ release}.
\end{quote}

\paragraph{Closed-loop wiping.}
The Spill wipe task demonstrates refinement of an existing
physical capability. In the five-seed development evaluation, performance
increased from \textbf{4/5 (80\%) in Round 4 to 5/5 (100\%) in Round 5}.

Earlier wiping behavior was largely open loop. A stroke could miss its
desired endpoint by several centimeters, while the agent continued without
checking which debris remained. The behavior also did not consistently
account for the offset between the fingertip and the actual wiping surface
of the gripper.

The improved behavior turns wiping into a closed-loop coverage procedure.
The robot estimates the support surface, calibrates contact height, executes
short overlapping strokes, retreats to obtain an unobstructed observation,
and selects subsequent strokes according to the remaining debris.

\begin{quote}
\small
\texttt{estimate patch $\rightarrow$ calibrate contact $\rightarrow$ short
pass $\rightarrow$ reobserve residuals $\rightarrow$ corrective pass
$\rightarrow$ verify clearance}.
\end{quote}

Thus, self-improvement does not merely tune a pushing trajectory; it turns
a brittle open-loop primitive into an observation-driven wiping procedure.

\paragraph{Verified bimanual lifting.}
The Two-arm lift task illustrates improvement in
coordinated bimanual manipulation. In the five-seed development evaluation,
success is \textbf{3/5 (60\%) in Round 11}, while later rounds reach
\textbf{5/5 (100\%)}.

Earlier behavior could treat the two arms almost independently: one arm
might miss its intended approach pose while the system nevertheless
continued toward grasping or lifting. Conversely, an ambiguous observation
could cause the robot to reset even after a plausible two-handed hold had
already been established.

The improved behavior introduces explicit bilateral verification. Each arm is assigned to the handle on its side, both lifting paths are checked before the grippers close, both grasps are verified, and the robot first performs a small synchronized test lift. It then continues with matched incremental motions while monitoring the object's rise and the relative heights of its two sides.

\begin{quote}
\small
\texttt{localize handles $\rightarrow$ check both paths $\rightarrow$ verify
both grasps $\rightarrow$ paired test lift $\rightarrow$ synchronized lift
$\rightarrow$ verify object rise}.
\end{quote}

These examples illustrate how self-improvement accumulates reusable procedural knowledge rather than merely adjusting isolated motion parameters. Across
rounds, the learned behaviors increasingly encode geometric relationships,
preconditions, verification steps, recovery strategies, and task-level
completion criteria.


\subsection{Round-by-Round Task Evolution}
\label{app:round_evolution}

Table~\ref{tab:round_task_evolution} reports per-task results
over all 15 rounds using the 10-seed held-out evaluation.
Each task entry reports the number of successful episodes out of 10.
These evaluations are separate from the five-rollout evaluations used inside
the self-improvement loop.
We report every round rather than only the final result because the evolution
is not necessarily monotonic: individual capabilities can emerge abruptly,
temporarily regress, and later recover.

Several patterns are visible.
Some tasks reach $10/10$ early and remain there, such as
Lift cube and Stack cubes.
Others show sharp improvements in later rounds.
For example, Nut assembly changes from $5/10$ in Round~4 to
$10/10$ in Round~5, while Place bottles in bin remains near zero for many
rounds before improving sharply in Round~12 and reaching $10/10$ by
Round~15.
Bimanual and long-horizon tasks exhibit more non-monotonic trajectories,
illustrating that self-improvement changes multiple reusable behaviors rather
than independently optimizing each benchmark instance.

\begin{table*}[t]
\centering
\caption{
Round-by-round task performance on ten held-out seeds per task.
Task entries are successes out of ten episodes.
D, V, and H denote development, validation, and feedback-held-out tasks,
respectively. Feedback-held-out tasks do not generate direct task-specific
improvement feedback but remain in cross-task regression validation.
The final row reports the overall success rate across the 22-task evaluation
($220$ episodes per round).
These numbers are \emph{not} the five-rollout in-loop candidate-evaluation
statistics used during self-improvement.
}
\label{tab:round_task_evolution}
\setlength{\tabcolsep}{3pt}
\renewcommand{\arraystretch}{1.15}
\resizebox{\textwidth}{!}{%
\begin{tabular}{llccccccccccccccc}
\toprule
Task & Split
& R1 & R2 & R3 & R4 & R5
& R6 & R7 & R8 & R9 & R10
& R11 & R12 & R13 & R14 & R15 \\
\midrule
Extract from fixture
& D & 2/10 & 5/10 & 5/10 & 10/10 & 10/10
& 10/10 & 9/10 & 10/10 & 9/10 & 10/10
& 9/10 & 10/10 & 8/10 & 10/10 & 10/10 \\

Insert into fixture
& D & 2/10 & 3/10 & 4/10 & 7/10 & 9/10
& 9/10 & 8/10 & 10/10 & 9/10 & 10/10
& 10/10 & 8/10 & 10/10 & 9/10 & 10/10 \\

Place dish on rack
& D & 0/10 & 0/10 & 0/10 & 2/10 & 3/10
& 6/10 & 5/10 & 5/10 & 8/10 & 4/10
& 9/10 & 8/10 & 9/10 & 9/10 & 10/10 \\

Serve onto plate
& H & 1/10 & 5/10 & 8/10 & 10/10 & 10/10
& 10/10 & 10/10 & 10/10 & 10/10 & 10/10
& 10/10 & 10/10 & 10/10 & 10/10 & 9/10 \\

Sort cubes (extended)
& D & 0/10 & 0/10 & 0/10 & 3/10 & 2/10
& 8/10 & 10/10 & 10/10 & 9/10 & 10/10
& 10/10 & 9/10 & 10/10 & 8/10 & 10/10 \\

Lift cube
& D & 10/10 & 10/10 & 10/10 & 10/10 & 10/10
& 10/10 & 10/10 & 10/10 & 10/10 & 10/10
& 10/10 & 10/10 & 10/10 & 10/10 & 10/10 \\

Nut assembly
& D & 0/10 & 0/10 & 0/10 & 5/10 & 10/10
& 3/10 & 6/10 & 6/10 & 3/10 & 5/10
& 5/10 & 4/10 & 9/10 & 10/10 & 9/10 \\

Restack cubes
& V & 4/10 & 6/10 & 6/10 & 10/10 & 10/10
& 10/10 & 10/10 & 10/10 & 10/10 & 10/10
& 10/10 & 9/10 & 10/10 & 9/10 & 10/10 \\

Spill wipe
& D & 0/10 & 0/10 & 0/10 & 6/10 & 10/10
& 10/10 & 10/10 & 10/10 & 10/10 & 10/10
& 10/10 & 10/10 & 10/10 & 10/10 & 10/10 \\

Stack cubes
& D & 9/10 & 9/10 & 10/10 & 10/10 & 10/10
& 10/10 & 10/10 & 10/10 & 10/10 & 10/10
& 10/10 & 10/10 & 10/10 & 10/10 & 10/10 \\

Two-arm handover
& H & 3/10 & 2/10 & 4/10 & 9/10 & 7/10
& 8/10 & 10/10 & 7/10 & 10/10 & 9/10
& 9/10 & 7/10 & 8/10 & 10/10 & 10/10 \\

Two-arm lift
& D & 0/10 & 2/10 & 2/10 & 9/10 & 9/10
& 2/10 & 2/10 & 4/10 & 4/10 & 6/10
& 6/10 & 4/10 & 10/10 & 8/10 & 10/10 \\

Fold towel
& D & 0/10 & 1/10 & 0/10 & 0/10 & 0/10
& 1/10 & 4/10 & 5/10 & 4/10 & 4/10
& 3/10 & 3/10 & 3/10 & 4/10 & 3/10 \\

Transfer object
& H & 4/10 & 6/10 & 8/10 & 10/10 & 10/10
& 10/10 & 10/10 & 10/10 & 10/10 & 10/10
& 10/10 & 10/10 & 10/10 & 10/10 & 9/10 \\

Lift rod
& D & 9/10 & 9/10 & 6/10 & 10/10 & 10/10
& 10/10 & 10/10 & 10/10 & 10/10 & 10/10
& 10/10 & 10/10 & 10/10 & 10/10 & 10/10 \\

Pick up cup
& D & 9/10 & 10/10 & 9/10 & 10/10 & 10/10
& 10/10 & 10/10 & 10/10 & 10/10 & 10/10
& 10/10 & 10/10 & 10/10 & 10/10 & 10/10 \\

Place bottles in bin
& D & 0/10 & 0/10 & 0/10 & 0/10 & 1/10
& 1/10 & 1/10 & 0/10 & 0/10 & 1/10
& 2/10 & 9/10 & 7/10 & 8/10 & 10/10 \\

Close sunglasses case
& H & 0/10 & 1/10 & 0/10 & 10/10 & 10/10
& 9/10 & 8/10 & 10/10 & 7/10 & 8/10
& 10/10 & 10/10 & 10/10 & 9/10 & 10/10 \\

Close drawer
& D & 4/10 & 3/10 & 8/10 & 10/10 & 10/10
& 10/10 & 10/10 & 10/10 & 10/10 & 10/10
& 10/10 & 10/10 & 10/10 & 10/10 & 10/10 \\

Sort cubes (standard)
& D & 1/10 & 1/10 & 2/10 & 4/10 & 5/10
& 8/10 & 10/10 & 9/10 & 10/10 & 9/10
& 9/10 & 10/10 & 8/10 & 10/10 & 9/10 \\

Stack blocks
& D & 2/10 & 3/10 & 4/10 & 7/10 & 6/10
& 6/10 & 8/10 & 7/10 & 10/10 & 10/10
& 9/10 & 10/10 & 10/10 & 7/10 & 10/10 \\

Transport cup
& D & 3/10 & 7/10 & 9/10 & 10/10 & 10/10
& 10/10 & 10/10 & 10/10 & 10/10 & 9/10
& 10/10 & 10/10 & 10/10 & 10/10 & 10/10 \\

\midrule
\textbf{Overall}
& &
29\% & 38\% & 43\% & 74\% & 78\%
& 78\% & 82\% & 83\% & 83\% & 84\%
& 87\% & 87\% & 92\% & 91\% & 95\% \\
\bottomrule
\end{tabular}%
}
\end{table*}

For Place bottles in bin, the Round~12 update additionally improved the
\emph{in-loop} five-rollout evaluation from $1/5$ to $5/5$.
The $9/10$ entry in Table~\ref{tab:round_task_evolution} is the separate
10-seed held-out evaluation of the resulting Round~12 system.


\subsection{Transfer without Direct Task-Specific Improvement Feedback}
\label{app:heldout_generalization}

A central question is whether reusable behaviors acquired through self-improvement transfer beyond the tasks that directly generate improvement feedback. We therefore track feedback-held-out tasks that are never analyzed by the Video Analyzer and never generate task-specific candidate revisions. These tasks remain in cross-task validation as regression checks, so the experiment measures transfer without direct task-specific improvement feedback rather than complete exclusion from development-time evaluation.

Table~\ref{tab:heldout_generalization} summarizes representative checkpoints
for tasks that do not provide direct task-specific improvement feedback.
The first four tasks belong to the main 22-task evaluation.
Grasp plate is an additional no-feedback task evaluated in its own
held-out evaluation grid and is therefore reported separately rather than pooled into the
main benchmark.

\begin{table*}[t]
\centering
\caption{
Performance on tasks without direct task-specific improvement feedback across
representative rounds under the 10-seed held-out evaluation. Failures on these
tasks are not analyzed by the Video Analyzer and do not generate candidate
revisions. The first four tasks remain in cross-task regression validation.
Grasp plate belongs to a separate no-feedback evaluation grid and is shown
here only as an additional transfer test.
}
\label{tab:heldout_generalization}
\setlength{\tabcolsep}{4.0pt}
\renewcommand{\arraystretch}{1.08}
\begin{tabular}{lccccccc}
\toprule
Task & R1 & R4 & R8 & R10 & R11 & R12 & R15 \\
\midrule
Serve onto plate
& 1/10 & 10/10 & 10/10 & 10/10 & 10/10 & 10/10 & 9/10 \\

Two-arm handover
& 3/10 & 9/10 & 7/10 & 9/10 & 9/10 & 7/10 & 10/10 \\

Transfer object
& 4/10 & 10/10 & 10/10 & 10/10 & 10/10 & 10/10 & 9/10 \\

Close sunglasses case
& 0/10 & 10/10 & 10/10 & 8/10 & 10/10 & 10/10 & 10/10 \\

\midrule
Grasp plate
& 2/10 & 7/10 & 9/10 & 10/10 & 10/10 & 6/10 & 6/10 \\
\bottomrule
\end{tabular}
\end{table*}

The results on these tasks provide evidence that the accumulated changes are
not restricted to the exact tasks that generated direct task-specific
improvement feedback.
For example, Two-arm handover improves from $3/10$ in
Round~1 to $10/10$ in Round~15, while Close sunglasses case improves
from $0/10$ to $10/10$.
Similarly, Transfer object reaches $10/10$ by Round~4 despite never generating
direct task-specific improvement feedback.

Transfer is not monotonic or universal.
The additional Grasp plate task reaches $10/10$ in Rounds~10--11
but later decreases to $6/10$.
We therefore interpret these results as evidence that reusable manipulation
changes can transfer to tasks without direct task-specific improvement feedback,
rather than as a claim that every library update uniformly improves every such
task.


\subsection{Failure and Learned-Correction Taxonomy}
\label{app:failure_taxonomy}

To characterize what is being accumulated during self-improvement, we organize
the failure analyses into recurring manipulation failure modes and the
corresponding procedural corrections.
This taxonomy is qualitative: it summarizes recurring patterns observed in the
analyzer outputs rather than reporting category frequencies.

\begin{table*}[t]
\centering
\caption{
Qualitative taxonomy of recurring manipulation failures and the reusable
corrections learned from them.
Many corrections introduce explicit
\emph{preconditions}, \emph{verification}, or \emph{closed-loop recovery},
rather than merely changing a Cartesian target.
}
\label{tab:failure_taxonomy}
\setlength{\tabcolsep}{5pt}
\renewcommand{\arraystretch}{1.15}
\begin{tabular}{
p{0.16\textwidth}
p{0.25\textwidth}
p{0.39\textwidth}
p{0.14\textwidth}}
\toprule
Failure mode
& Typical failure
& Learned correction
& Representative tasks \\
\midrule

\textbf{Localization and geometry uncertainty}
&
The policy acts on a stale, ambiguous, or poorly localized object,
receptacle, or support surface.
&
Re-observe after scene changes; reconstruct geometry from multiple visible
points and depth; invalidate stale targets after contact or failed motion.
&
Place bottles in bin,
Insert into fixture,
Place dish on rack
\\

\textbf{Grasp identity and retention}
&
A blocked or object-sized gripper opening is treated as proof that the desired
object is held, even when the grasp is empty, mixed, or on the wrong object.
&
Combine gripper state with wrist/top observations and a short test lift;
require visible target co-motion before transport; preserve uncertain grasps
rather than opening blindly.
&
Extract from fixture,
Nut assembly,
Stack blocks
\\

\textbf{Incomplete path feasibility}
&
An endpoint is reachable, but an approach, lift, transit, release, or retreat
waypoint along the actual manipulation route is not.
&
Check the complete manipulation path before committing, including approach,
grasp, lift, transport, release, and retreat, using the actual arm configuration and yaw.
&
Place bottles in bin,
Place dish on rack,
Stack blocks
\\

\textbf{End-effector-centric geometry}
&
The fingertip reaches the nominal target while the relevant object feature
does not: e.g., the bottle misses the opening or the nut hole misses the peg.
&
Track the geometry of the manipulated object and its task-relevant feature;
estimate object-to-gripper offsets and align the object, hole, bottom surface,
or footprint rather than the fingertip.
&
Place bottles in bin,
Nut assembly,
Insert into fixture
\\

\textbf{Contact and support calibration}
&
Motion is commanded relative to the fingertip without accounting for pad
extension, object extent, table height, or support geometry, causing
penetration or ineffective contact.
&
Estimate the support surface and relevant tool/object offset; use
surface-relative heights and small monitored increments instead of large
open-loop contact motions.
&
Spill wipe,
Fold towel
\\

\textbf{Multi-effector coordination}
&
Each arm is controlled locally without verifying that both contacts are valid
or that the required relation between the two arms is maintained.
&
Verify both grasps, perform a paired test motion, preserve relative geometry,
and use synchronized incremental motions with an explicit relation tolerance.
&
Two-arm lift
\\

\textbf{Execution-budget failure}
&
A physically valid strategy does not finish because repeated staging,
recovery, or redundant motion consumes the remaining robot-time budget.
&
Plan complete routes using measured execution time; resolve difficult
cross-workspace transfers early; reserve explicit time for the final
delivery and verification.
&
Sort cubes (extended),
Sort cubes (standard)
\\

\textbf{Premature release or completion}
&
Successful execution of a primitive is treated as task success even though
the object is not seated, contained, supported, or geometrically correct.
&
Make irreversible actions fail closed; verify physical task state after
release; preserve a held object under uncertainty; trigger bounded recovery
when the final state is not confirmed.
&
Place bottles in bin,
Stack blocks,
Fold towel
\\

\bottomrule
\end{tabular}
\end{table*}

Across these categories, a common trend emerges.
Self-improvement progressively converts implicit assumptions into explicit
procedural checks.
A nominally successful motion is no longer sufficient evidence of a successful
grasp, insertion, placement, or coordinated manipulation.
Instead, the evolved procedures increasingly couple action with observation:
they verify preconditions before acting, monitor execution while acting, and
check task-level state before terminating.

This perspective also clarifies why the learned changes can transfer between
tasks.
Corrections such as test-lift verification, full-route reach checking,
object-centric alignment, incremental contact motion, and post-release
verification are not tied to a single object instance.
They are reusable manipulation patterns that can be composed in later tasks.


\subsection{Limitation Case: Deformable Manipulation}
\label{app:fold_towel_limitation}

Self-improvement does not solve all manipulation regimes.
The clearest remaining limitation is deformable-object manipulation, as
illustrated by Fold towel.

Table~\ref{tab:fold_towel_evolution} reports results on ten held-out seeds per round.
Unlike many rigid-object tasks, folding never reaches reliable performance.
The standard task peaks at $5/10$ successes, while the stricter
Fold towel (hard) variant remains at $0/10$ throughout all 15 rounds.
The hard variant uses the same seeded initial configurations but tightens the
corner-alignment tolerance from 5\,cm to 1\,cm.

\begin{table}[t]
\centering
\caption{
Towel-folding performance on ten held-out seeds across self-improvement rounds.
Fold towel (hard) tightens the required corner alignment from
5\,cm to 1\,cm and is evaluated separately from the main 22-task benchmark.
}
\label{tab:fold_towel_evolution}
\setlength{\tabcolsep}{2.7pt}
\renewcommand{\arraystretch}{1.08}
\resizebox{\columnwidth}{!}{%
\begin{tabular}{lccccccccccccccc}
\toprule
& R1 & R2 & R3 & R4 & R5
& R6 & R7 & R8 & R9 & R10
& R11 & R12 & R13 & R14 & R15 \\
\midrule

Fold towel
& 0/10 & 1/10 & 0/10 & 0/10 & 0/10
& 1/10 & 4/10 & 5/10 & 4/10 & 4/10
& 3/10 & 3/10 & 3/10 & 4/10 & 3/10 \\

Fold towel (hard)
& 0/10 & 0/10 & 0/10 & 0/10 & 0/10
& 0/10 & 0/10 & 0/10 & 0/10 & 0/10
& 0/10 & 0/10 & 0/10 & 0/10 & 0/10 \\

\bottomrule
\end{tabular}%
}
\end{table}

Failure analysis suggests that this limitation is not explained by a single
missing waypoint.
Successful folding requires maintaining thin, deformable material in both
grippers while the geometry of the cloth changes throughout the motion.
Observed failures include uncertain corner acquisition, loss of one corner
during bimanual transport, insufficient support clearance, incomplete landing
reachability, and termination after a visually plausible but geometrically
incorrect partial fold.

Several of the corrections learned for rigid-object manipulation remain useful:
the robot can verify a corner with a short lift, check the complete motion path,
move the two arms in shorter matched increments, and re-observe the cloth after
a slip.
However, these mechanisms still do not provide sufficiently reliable
estimation of layer identity, persistent cloth contact, and post-fold geometry.

This result exposes an important boundary of the current approach.
For rigid objects, many failures can be converted into reusable procedural
guards around grasping, reachability, alignment, release, and verification.
For deformable objects, the underlying state itself changes substantially under
contact, and visually similar configurations can correspond to different
layering and attachment states.
Improving this regime likely requires richer representations of deformable
state and contact, rather than only additional composition of the existing
rigid-object skills.

\subsection{Simulation Task Suite}
\label{app:simulation_tasks}

We evaluate RPG on a suite of 22 bimanual manipulation tasks on the YAM
station. Each task specification fixes the simulation scene, language
instruction, execution budget, and a binary evaluator that evaluates the final
outcome from simulator state. The policy's own success claims are recorded for
diagnostics but never affect the success label.

The suite contains 17 development tasks, one validation task, and four feedback-held-out tasks. Development tasks provide task-specific failure feedback and may generate candidate revisions. The validation task, Restack cubes, does not generate candidate revisions and is used as a cross-task regression check. It is reported separately from the feedback-held-out tasks because only the latter are used in the transfer analysis of Appendix~\ref{app:heldout_generalization}. The four feedback-held-out tasks are Serve onto plate, Two-arm handover, Transfer object, and Close sunglasses case. They are neither analyzed by the Video Analyzer nor used to generate task-specific candidate revisions. They remain in cross-task validation only as regression checks.

All tasks use the same observation interface: RGB images from the top and both
wrist cameras, top-camera depth, and robot proprioception. Initial conditions
randomize object poses and selected physical parameters, including friction,
mass, table height, camera pose, and depth noise, followed by a fixed settling
period before execution. Each trial consists of one episode. Task-specific limits on skill calls and robot
execution time are part of the task definition. Unless otherwise noted, task success is scored from the final simulator state. RATs follows its native any-time success criterion as reported in Table~\ref{tab:per-task}.

\paragraph{Outcome-based evaluators.}
The benchmark uses nine evaluator types that capture different manipulation
objectives. Most placement tasks use an \texttt{all\_in\_targets} evaluator,
which checks that every required object lies within task-specific horizontal
and vertical tolerances around its target. Other evaluators check lifting,
articulated closure, extraction, stacking, restacking, level bimanual lifting,
debris removal, and cloth folding. Several evaluators include additional guard
conditions to reject degenerate solutions. For example, moving an entire
fixture does not satisfy extraction, and a pre-stacked scene does not satisfy
restacking unless the base itself is relocated.

\paragraph{Representative task definitions.}
For Insert into fixture, both pegs must reach their corresponding cups, with
a $3$\,cm horizontal tolerance. For Two-arm lift, both
handle sites must rise by at least $6$\,cm while their rise differs by less
than $3$\,cm, preventing a single-arm tilted lift from passing.
For Spill wipe, at least six of nine debris particles must
finish outside the marked patch, so merely sweeping one portion of the patch
is insufficient.

The Place bottles in bin scene contains six bottles and a central bin, but
only two specified bottles are targets: the black bottle and the clear bottle
with the blue-and-yellow label. The other four are distractors and need not be
moved. Towel folding is evaluated directly from the cloth mesh: corresponding
corners must align within $5$\,cm, the moved layer must remain above the
receiving layer, the projected footprint must shrink to $42$--$58\%$ of its
initial area, and the resulting cloth must remain sufficiently thin.

\begin{table}[t]
\centering
\caption{Composition of the 22-task simulation benchmark.}
\label{tab:task_suite_summary}
\setlength{\tabcolsep}{5pt}
\renewcommand{\arraystretch}{1.08}
\begin{tabular}{lcc}
\toprule
Split & \# Tasks & Examples \\
\midrule
Development
& 17
& insertion, stacking, wiping, two-arm lift \\
Validation
& 1
& restack cubes \\
Feedback-held-out
& 4
& serving, handover, case closing \\
\bottomrule
\end{tabular}
\end{table}

Two additional diagnostic tasks are evaluated outside the main 22-task grid.
Fold towel (hard) uses the same folding task with the corner
tolerance tightened from $5$\,cm to $1$\,cm, while
Grasp plate isolates rim grasping. Neither is pooled into the main
benchmark.

\end{document}